\documentclass[a4paper,conference]{IEEEtran}
\IEEEoverridecommandlockouts
\usepackage{amsmath,amssymb,amsfonts}
\usepackage{algorithmic}
\usepackage{graphicx}
\usepackage{textcomp}
\usepackage{xcolor}
\usepackage{hyperref}
\usepackage{booktabs}
\usepackage[caption=false,font=footnotesize]{subfig}
\hypersetup{
  colorlinks   = true,    
  urlcolor     = blue,    
  linkcolor    = blue,    
  citecolor    = green    
}

\usepackage[backend=biber]{biblatex}
\graphicspath{{figs/}}

\begin{document}

\title{Video Anomaly Detection by Estimating Likelihood of Representations}

\author{\IEEEauthorblockN{Yuqi Ouyang}
\IEEEauthorblockA{Department of Computer Science \\
University of Warwick\\
Coventry, UK \\
yuqi.ouyang@warwick.ac.uk}
\and
\IEEEauthorblockN{Victor Sanchez}
\IEEEauthorblockA{Department of Computer Science \\
University of Warwick\\
Coventry, UK \\
v.f.sanchez-silva@warwick.ac.uk}
}

\maketitle

\begin{abstract}
Video anomaly detection is a challenging task not only because it involves solving many sub-tasks such as motion representation, object localization and action recognition, but also because it is commonly considered as an unsupervised learning problem that involves detecting outliers. Traditionally, solutions to this task have focused on the mapping between video frames and their low-dimensional features, while ignoring the spatial connections of those features. Recent solutions focus on analyzing these spatial connections by using hard clustering techniques, such as K-Means, or applying neural networks to map latent features to a general understanding, such as action attributes. In order to solve video anomaly in the latent feature space, we propose a deep probabilistic model to transfer this task into a density estimation problem where latent manifolds are generated by a deep denoising autoencoder and clustered by expectation maximization. Evaluations on several benchmarks datasets show the strengths of our model, achieving outstanding performance on challenging datasets.
\end{abstract}


\section{Introduction}
Video anomaly detection refers to detecting abnormal activities or events in a scene. This task is closely related to action recognition as both can be solved by activity classification based on extracted appearance and motion features. Since it is very challenging to collect and label examples of all possible types of abnormal events in a scene, video anomaly detection is usually solved as an unsupervised learning task, where the objective is to detect outliers.

Video anomaly detection can be addressed by learning to model spatio-temporal features extracted from normal video data. The learned model can then be used to detect abnormal events by determining how well a given video fits the model. Early work based on this strategy uses hand-crafted features, e.g., Histograms of Oriented Gradients (HOG) \cite{HOG} or Histograms of Oriented Flows (HOF) \cite{HOF}, and dictionary learning to build a model \cite{150FPS,Roberto}. 

Recently, deep learning has been shown to attain an excellent performance on many computer vision tasks including image classification \cite{AlexNet}, object detection \cite{YOLO}, action recognition \cite{TwoStream}, and anomaly detection \cite{GANNB,LSA}. Notably, autoencoders (AEs) have been used to detect anomalies in videos. An AE comprises an encoder and a decoder network that are trained to minimize the difference between the input data and the reconstructed output data. AEs can then be used to detect abnormal patterns by measuring the Mean Square Error (MSE) between the input and output data; e.g., if the MSE is higher than a threshold value, then a video is deemed to be abnormal. In this context, the threshold value is commonly set by analyzing the MSE values of a set of normal videos. Unfortunately, it is possible for AEs to reconstruct normal video data with high MSE values \cite{MemAE}. This shows that the MSE may not be an accurate metric for statistically detecting video abnormalities. 

\begin{figure*}[!ht]
\centering
\includegraphics[scale=0.275]{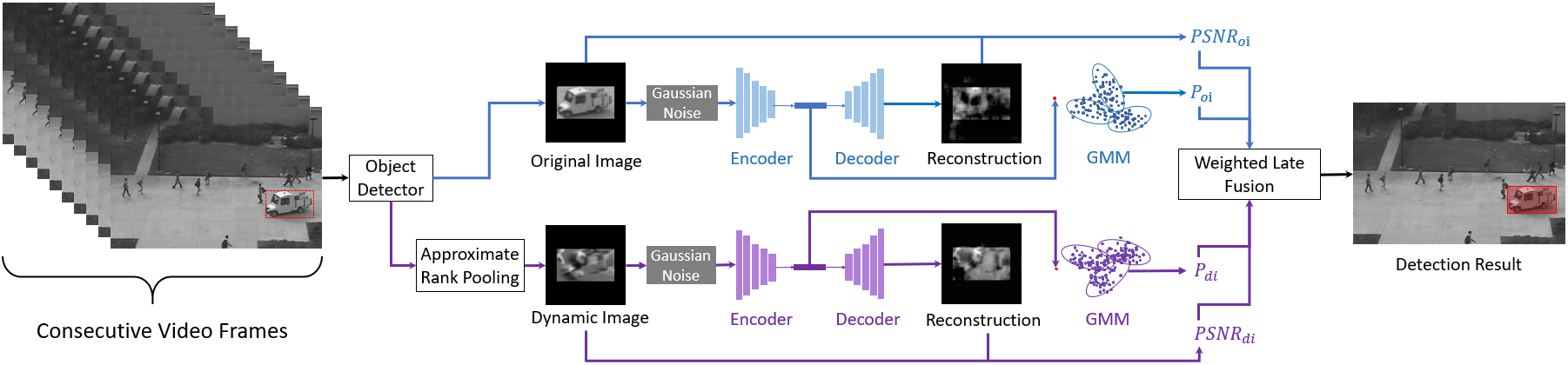}
\caption{The proposed GMM-DAE model. An object detector is initially applied to consecutive video frames to generate patches. A dynamic image is then computed for each patch by approximate rank pooling. Two denoising autoencoders are trained to reconstruct the patches and their corresponding dynamic images. Low-dimensional manifolds are clustered with two Gaussian Mixture Models (GMM). The final anomaly score is computed by fusing the reconstruction errors ($PSNR_{oi}$ and $PSNR_{di}$) and latent likelihood ($P_{oi}$ and $P_{di}$) of the appearance (in blue) and motion (in purple) pipelines.}
\label{fig:ProjectFlow}
\end{figure*}

Low-dimensional features generated from an encoding process have been shown to accurately reflect the variances of the input data in the feature space \cite{Hinton}. This characteristic can then be exploited to distinguish normal and abnormal videos. For example, one can measure the differences between the low-dimensional representations of normal and abnormal video data \cite{MemAE,OCA,LSA,DGK,MLEP}. Alternatively, anomaly detection can be solved as a density estimation task by using the distribution of the latent manifolds generated by normal data to distinguish abnormal data \cite{DAGMM}. Inspired by this strategy and the two-stream model widely used for action recognition \cite{TwoStream}, we propose to use soft clustering on low-dimensional features to detect video abnormalities. Specifically, we introduce a simple but effective deep probabilistic model named GMM-DAE. Our model applies Expectation Maximization (EM) to train a Gaussian Mixture Model (GMM) to estimate the data distribution in the feature space, where latent manifolds are generated by the encoder of a deep denoising autoencoder (DAE). The overall structure of our model is illustrated in Fig.~\ref{fig:ProjectFlow}. 

The main contributions of this work are summarized next:
\begin{itemize}
\item Soft-clustering on deep features is applied for video anomaly detection for the first time. Although the work in \cite{UCSD} also attempts to solve this task by using GMMs, GMM-DAE differs from that model in two crucial aspects: first, approximate rank pooling is used to more accurately capture motion information; second, a deep denoising autoencoder is used in lieu of principal component analysis (PCA) to more robustly learn spatio-temporal representations.
\item A thorough performance analysis of GMM-DAE is provided to show the potential of density estimation models for video anomaly detection.
\item Compared with recent baseline models, GMM-DAE attains a very competitive performance on the UCSD Ped2 and CUHK Avenue datasets, while achieving the State-Of-The-Art (SOTA) performance on the challenging ShanghaiTech dataset.
\end{itemize}

We organize the rest of this paper as follows. Section \ref{RelatedWork} briefly reviews related work. Section \ref{ModelDetails} presents the details of our model. Section \ref{Experiments} presents and discusses the evaluation results. Finally, Section \ref{Conclusion} concludes this paper.

\section{Related Work} \label{RelatedWork}
This section focuses on recent works on video anomaly detection using deep learning. We divide these works into two categories: those based on \emph{reconstruction and prediction} and those based on \emph{memorization and density estimation}.

\noindent \textbf{Reconstruction and Prediction.}
These methods use an encoder-decoder neural network to either reconstruct the video or predict future frames from the latent representations. They are based on the MSE between the generated and desired outputs. The general idea is that the MSE is expected to be high if the input video contains abnormal patterns. There are two main types of networks used by these works to extract spatio-temporal features from the input: 3D Convolutional AEs (3DConv-AEs) and combined structures formed by a Convolutional Neural Networks (CNNs) and Recurrent Neural Network (RNNs), e.g., a CNN with a Long Short-Term Memory (LSTM) designed as an AE (ConvLSTM-AE). Researchers have used 3DConv-AEs for video anomaly detection by measuring the MSE of the reconstructed output \cite{3DConvLSTMR,3DConv2StreamR} or the MSE of several predicted frames \cite{3DConvRP}. ConvLSTM-AEs have also been used for video anomaly detection in the same manner \cite{3DConvLSTMR,ConvLSTMRP}. Although these methods have been shown to attain promising performances, their training times are usually very long.

\noindent \textbf{Memorization and Density Estimation.}
Video anomaly detection by memorization refers to initially learning a collection of normal latent representations, and then detecting abnormalities by determining if a new latent representation fits within the learned collection. Learning the distribution of the training video data or the distribution of their latent representations can also be regarded as a memory-based technique. In \cite{Xu}, a two-steam method is proposed to detected abnormal video data by applying multiple one-class support vector machines to learn the latent feature collections of training video frames. The work in \cite{DGK} maps latent representations into different scores of action attributes and records these action attributes during training; the abnormal events are then detected by the action attributes they trigger. A memory-augmented deep AE is designed in \cite{MemAE} for dictionary learning. This method updates a weight matrix during training with normal videos. By searching the feature memory, this weight matrix is then applied to generate a normal output most similar to the input, thus the model differs from the vanilla AE by amplifying the MSE between the produced and desired outputs.

The research community has also seen an increasing number of deep probabilistic models for video anomaly detection. One could detect anomalies by identifying those videos that do not fit the estimated distribution of normal videos. Generative Adversarial Networks (GANs) is an appropriate option for this as GANs are designed based on the idea of distribution estimation. GANs have been used to detect anomalies in medical image data \cite{MedicalGAN}, security image data \cite{SecurityGAN} and video data \cite{GANNB,GANOS,GANROIOF,GANMLP}. However, these GAN-based methods have not achieved strong performances.

It has been shown that by minimizing the MSE between the output data and the ground truth while maximizing the likelihood of the generated latent representations, the probability density of the training video data can be estimated \cite{LSA,DAGMM}. Anomalies can then be detected by measuring how well the latent representations fit the distribution estimated in the training stage, where the data likelihood is commonly used as the fitting metric. Our GMM-DAE model is based on this principle.

\section{Proposed Denoising Autoencoder with Expectation-Maximization} \label{ModelDetails}
Our model comprises four parts: image patch generation, encoding/decoding via two deep DAEs, density estimation and anomaly inference. As illustrated in Fig.~\ref{fig:ProjectFlow}, two deep DAEs are used to generate latent representations for the patches extracted from the frames (appearance) and their dynamic images (motion). Each latent manifold is then used separately for anomaly inference based on its likelihood value using a trained GMM. Peak Signal-to-Noise Ratio (PSNR) values of the reconstructed data are also used for anomaly inference. Finally, late fusion is applied to detect abnormalities, in terms of appearance and motion, by voting.

\subsection{Image Patch Generation}
Based on its detection speed and accuracy, YOLOv3 \cite{YOLOv3} is applied to extract patches from the current frame. For each detected object (i.e., patch) in the current frame, we use its bounding box to crop consecutive image patches from previous frames, thus we do not track objects. The motion information of an object is summarized into a dynamic image by approximate rank pooling \cite{DynaImg}. Dynamic images, like optical flow \cite{PWCNet}, can also represent an object's motion but with a lower computational complexity. Specifically, given a set of $t$ patches, $\{x^1, x^2, ..., x^t\}$, for a time stride parameter, $t$, the dynamic image, $d^t$, with respect to the current patch $x^t$ is calculated as:

\begin{equation}
\centering
d^t = \sum_{i = 1}^{t} \sum_{j = i}^{t} \left( \frac{2j -t -1}{j} \right) x^i.
\label{eq:DynaImg}
\end{equation}

For each detected object in the current frame, we then generate a patch in the pixel domain and a corresponding dynamic image. These are resized to a size of $64\times64$.

\subsection{Deep Denoising Autoencoder}
\begin{figure}[!htb]
\centering
\includegraphics[scale=0.33]{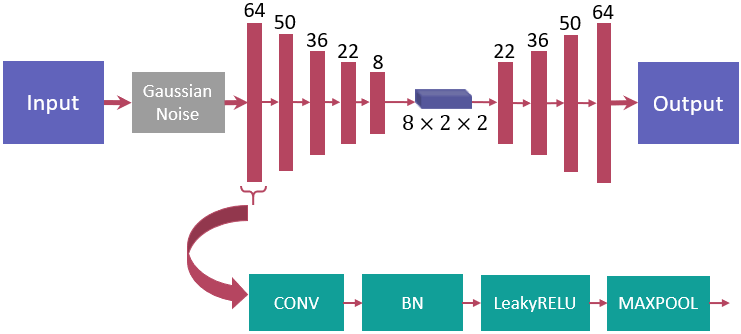}
\caption{Architecture of the DAE in the GMM-DAE model. CONV: convolutional layer. BN: batch normalization.}
\label{fig:AE}
\end{figure}

A DAE maps a corrupted input data, $\tilde{x}$, to a latent representation, $z$, by encoding it as $z = f_e(\tilde{x}; \theta_e)$ with parameters $\theta_e$. The generated latent representation, $z$, is reconstructed by decoding it as $\hat{x} = f_d(z; \theta_d)$ with parameters $\theta_d$. The original input data, $x$, is corrupted by adding a Gaussian noise, indicated as $\tilde{x}\sim{q_{noise}\left(\tilde{x}|{x}\right)}$ \cite{DAE,SDAE}, where $q_{noise}$ is an arbitrary noise distribution. We use isotropic Gaussian noise in our model as $\tilde{x}|{x}\sim{N\left(x,\sigma^2\mathcal{I}\right)}$, where $\sigma$ is the standard deviation and $\mathcal{I}$ is the identity matrix. The architecture of the DAE used in GMM-DAE is shown in Fig.~\ref{fig:AE}. 

All convolutional layers of the encoder have a filter size of $3\times3$, stride size of $1$ and zero-padding with a size of $1$. Each convolutional layer, except for the last one, uses LeakyReLU activation and is followed by batch normalization and a max-pooling operation with a kernel size of $2\times2$ and a stride size of $2$. The last convolutional layer of the encoder uses Sigmoid activation. Since it has been shown that batch normalization is not certainly connected to the internal covariate shift of the input for the next layer \cite{BatchNorm}, we widely use batch normalization in the encoder for stabilized training. The number of filters of each convolutional layer gradually decreases from 64 to 8. The output of the encoder has dimensions of $8\times2\times2$.

The decoder comprises transposed convolutional layers and max-unpooling layers to perform the opposite operations to the encoder. The number of filters of the transposed convolutional layers gradually increases from $22$ to $64$, where the last transposed convolutional layer has filters of size $1$ to reduce the dimensions from $64\times64\times64$ to $64\times64$.

Each DAE takes a $64\times64$ input $x$, and outputs $\hat{x}$ with the same size. Given a batch size of $n$ training examples, training aims at finding the parameter sets, $\theta_e$ and $\theta_d$, that minimize the following loss function based on $\ell_2$-norm:

\begin{equation}
\centering
L\left( \boldsymbol{X}, \boldsymbol{\hat{X}} \right) = 
\frac{1}{n} \sum_{i = 1}^{n} \lVert x_i-\hat{x}_i \rVert^2_2 +
\beta \lVert \boldsymbol{W} \rVert^2_2,
\label{eq:LF}
\end{equation}

\noindent where $\boldsymbol{{X}}$ and $\boldsymbol{\hat{X}}$ are the whole set of $n$ uncorrupted inputs and reconstructed outputs, respectively, and $\lVert \boldsymbol{W} \rVert^2_2$ denotes $\ell_2$-regularization on the parameters of the DAE by a factor $\beta$.

\subsection{Density Estimation}
Training each DAE with a set of normal inputs, $\boldsymbol{X}$, produces a set of latent representations, $\boldsymbol{Z}$. Such a set of latent representations represents the understanding of the normal training data, which can be statistically quantified by estimating its distribution. In the test stage, one can measure how well the generated latent representation, $z$, of a test input fits within the probability distribution of $\boldsymbol{Z}$. This measured difference is the basis to detect abnormalities. Expectation Maximization with a Gaussian Mixture Model (EMGMM) is a powerful technique to perform probability density estimation of data samples. Based on Maximum Likelihood Estimation (MLE), EM focuses on clustering samples by iteratively increasing their likelihood values with respect to the GMM \cite{EM}. The EM process is divided into three operations: GMM initialization, E-step and M-step, and convergence.

\noindent \textbf{GMM Initialization}. EMGMM is sensitive to the initialization of the GMM as poor initialization could lead to convergence to a bad local maximum. K-Means++ is a simple but effective method for initializing data centroids \cite{KMPP}. Hence, we first apply K-Means++ to cluster the latent manifold set, $\boldsymbol{Z}$, into $k$ clusters with centroids $\{ c_1, c_2, ... , c_k \}$. The clustered set $\boldsymbol{Z}$ is denoted as a collection of $k$ subsets $\{ \boldsymbol{z_1}, \boldsymbol{z_2}, \boldsymbol{...} , \boldsymbol{z_k} \}$. $k$ multivariate Gaussian blobs, $\{ N(\phi_1, \mu_1, \Sigma_1), N(\phi_2, \mu_2, \Sigma_2), ..., N(\phi_k, \mu_k, \Sigma_k) \}$ are then initialized, where the $j^{th}$ blob is initialized as follows:

\begin{equation}
\centering
\phi_j = \frac{D(\boldsymbol{z_j})}{\sum_{i = 1}^{k} D(\boldsymbol{z_i})},
\label{eq:InitPhi}
\end{equation}

\begin{equation}
\centering
\mu_j = \frac{1}{D(\boldsymbol{z_j})} \sum_{z \in \boldsymbol{z_j}} z,
\label{eq:IninMu}
\end{equation}

\begin{equation}
\centering
\Sigma_j = \frac{1}{D(\boldsymbol{z_j})} \sum_{z \in \boldsymbol{z_j}} (z-\mu_j)(z-\mu_j)^T,
\label{eq:InitSigma}
\end{equation}

\noindent where function $D(\cdot)$ returns the number of samples in a cluster, and $\phi_j$, $\mu_j$ and $\Sigma_j$ are the prior knowledge, the mean vector and the covariance matrix of the $j^{th}$ multivariate Gaussian blob, respectively. This mixture of $k$ Gaussian blobs are iteratively updated, as explained next.

\noindent \textbf{E-step and M-step}. EM applies the E-step to initially fit the current Gaussian blobs to samples and retrieve the posterior likelihood values. It then applies the M-step to update the parameters of the Gaussian blobs based on the posterior knowledge produced by the E-step. Likelihood values of samples are expected to increase by iteratively applying the E-step and M-step. Thus, EM estimates the data distribution by iteratively running the E-step and M-step. After converging, soft clusters are produced where a sample is assigned to each cluster by a likelihood value. 

In the E-step, the posterior likelihood value of sample $z_i$ for the $j^{th}$ Gaussian blob with parameters $N(\phi_j, \mu_j, \Sigma_j)$ is given by:

\begin{equation}
\centering
\gamma_{ij} = \frac{\omega_{ij} \cdot \phi_j \cdot \exp{-\frac{1}{2} \left( z_i - \mu_j \right)^T \Sigma_j^{-1} \left( z_i - \mu_j \right)}}{\sqrt{\lvert 2 \pi \Sigma_j \rvert}},
\label{eq:Estep}
\end{equation}

\noindent where $\omega_{ij}$ represents a normalization factor and $\lvert \ \cdot \ \rvert$ denotes the matrix determinant. 

In the M-step, the objective is to adjust the parameters of the Gaussian blobs based on the calculated posterior likelihoods, $\gamma_{ij}$. The updated parameters of the $j^{th}$ Gaussian blob after the M-step, denoted by $N(\hat{\phi}_j,\hat{\mu}_j, \hat{\Sigma}_j)$, are computed as follows:

\begin{equation}
\centering
\hat{\phi}_j = \sum_{i = 1}^{n} \frac{\gamma_{ij}}{n},
\label{eq:MstepPhi}
\end{equation}

\begin{equation}
\centering
\hat{\mu}_j = \frac{\sum_{i = 1}^{n} \gamma_{ij} z_i} {\sum_{i = 1}^{n} \gamma_{ij}},
\label{eq:MstepMu}
\end{equation}

\begin{equation}
\centering
\hat{\Sigma}_j = \frac{\sum_{i = 1}^{n} \gamma_{ij} \left( z_i - \hat{\mu}_j \right) \left( z_i - \hat{\mu}_j \right)^T} {\sum_{i = 1}^{n} \gamma_{ij}},
\label{eq:MstepSigma}
\end{equation}

\noindent where $n$ is the total number of training samples. As a result, $k$ multivariate Gaussian blobs are updated in the M-step.

\noindent \textbf{Convergence of EMGMM}. The increase in likelihood values for the whole set $\boldsymbol{Z}$ before and after the M-step can be calculated as $\Delta L(\boldsymbol{Z}) = \hat{L}(\boldsymbol{Z}) - L(\boldsymbol{Z})$, with $\hat{L}(\boldsymbol{Z})$ and $L(\boldsymbol{Z})$ computed as:

\begin{equation}
\centering
L(\boldsymbol{Z}) = \sum_{i = 1}^{n} \log \sum_{j = 1}^{k} \frac{\phi_j \cdot \exp{-\frac{1}{2} \left( z_i - \mu_j \right)^T \Sigma_j^{-1} \left( z_i - \mu_j \right)}}{\sqrt{\lvert 2 \pi \Sigma_j \rvert}},
\label{eq:EnergyBM}
\end{equation}

\begin{equation}
\centering
\hat{L}(\boldsymbol{Z}) = \sum_{i = 1}^{n} \log \sum_{j = 1}^{k} \frac{\hat{\phi}_j \cdot \exp{-\frac{1}{2} \left( z_i - \hat{\mu}_j \right)^T \hat{\Sigma}_j^{-1} \left( z_i - \hat{\mu}_j \right)}}{\sqrt{\lvert 2 \pi \hat{\Sigma}_j \rvert}}.
\label{eq:EnergyAM}
\end{equation}

The EMGMM process is said to converge if $\Delta L(\boldsymbol{z}) < \epsilon$, where $\epsilon$ is a small constant.

\subsection{Anomaly Inference}
For each data sample, $x$, the trained denoising autoencoder generates both its latent representation, $z$, and the reconstructed data sample, $\hat{x}$. We compute the PSNR, $PSNR(x,\hat{x})$, and the latent likelihood, as follows:

\begin{equation}
\centering
PSNR(x,\hat{x}) = 10\log_{10}{\frac{max{(x)}}{MSE(x,\hat{x})}}
\label{eq:PSNR}
\end{equation}

\begin{equation}
\centering
P(z) = \log \sum_{j = 1}^{k} \frac{\hat{\phi}_j \cdot \exp{-\frac{1}{2} \left( z - \hat{\mu}_j \right)^T \hat{\Sigma}_j^{-1} \left( z - \hat{\mu}_j \right)}}{\sqrt{\lvert 2 \pi \hat{\Sigma}_j \rvert}},
\label{eq:LL}
\end{equation}

\noindent where $max(x)$ is the maximum possible value of input $x$, and $MSE(x,\hat{x})$ is the MSE between the original input, $x$, and its reconstructed version, $\hat{x}$. The Gaussian blob set $N(\boldsymbol{\hat{\phi}}, \boldsymbol{\hat{\mu}}, \boldsymbol{\hat{\Sigma}})$, comprising $k$ multivariate Gaussian blobs and trained by the EM process, is used in (\ref{eq:LL}). As shown in Fig.~\ref{fig:ProjectFlow}, two different pipelines are designed for appearance and motion information analysis. Under the assumption that abnormalities result in low PSNR and latent likelihood values, we use late fusion to calculate the anomaly value, $A(x^t)$, of the current patch, $x^t$, as follows:

\begin{equation}
\centering
\begin{split}
A(x^t) & = - \left[ \lambda_{1} P(z_{x^t}) + \lambda_{2} \cdot PSNR(x^t,\hat{x}^{t}) \right. \\
& \left. + \lambda_{3} P(z_{d^t}) + \lambda_{4} \cdot PSNR(d^t,\hat{d}^{t}) \right],
\end{split}
\label{eq:AnoValByFusion}
\end{equation}

\noindent where $d^t$ indicates the dynamic image of current patch $x^t$. $z_{x^t}$ and $z_{d^t}$, respectively, denote the latent representations generated by the two DAEs. $\hat{x}^{t}$ and $\hat{d}^{t}$, respectively, represent the reconstructed outputs with inputs $x^t$ and $d^t$. $\lambda_{1}$, $\lambda_{2}$, $\lambda_{3}$ and $\lambda_{4}$ are user-defined hyper-parameters. To perform frame-level anomaly analysis for the current video frame, $I^t$, with $n$ detected objects (i.e., patches) $\{{x_{1}^{t}}, {x_{2}^{t}}, ...,{x_{n}^{t}}\}$, we define its anomaly value, $A(I^t)$, as the maximum anomaly value among its $n$ patches, namely:

\begin{equation}
\centering
A(I^t) = max\{ A(x_{1}^{t}), A(x_{2}^{t}), ...,A(x_{n}^{t}) \}. 
\label{eq:FrameAnoValByMax}
\end{equation}

\section{Experiments} \label{Experiments}
We evaluate GMM-DAE on three different datasets: UCSD Ped2, CUHK Avenue and ShanghaiTech. 

The UCSD dataset \cite{UCSD} focuses on abnormalities in a pedestrian walkway. We focus on the UCSD Ped2 dataset because the UCSD Ped1 dataset comprises videos with a low resolution, which do not reflect the video quality of current surveillance cameras. UCSD Ped2 comprises 2550 training frames and 2010 test frames, all with a size of $240\times360$.

The CUHK Avenue dataset \cite{150FPS} comprises one scene captured by a camera looking at an avenue near a subway entrance from a nearly horizontal view angle. It is a relatively challenging dataset due to the complex background and the variations in the objects' size induced by their distance to the camera. As done in \cite{DGK,OCA}, we exclude five test videos with missing annotations. The Avenue dataset contains 15328 frames for training and 10622 frames for testing, all with a size of $360\times640$.

The ShanghaiTech dataset \cite{GANNB} includes 13 different scenes, which makes it particularly challenging. This dataset has frames with a size of $480\times856$. It contains many complex types of abnormalities such as people fighting, pushing strollers, and riding a motorcycle, which are not included in the UCSD Ped2 and CUHK Avenue datasets. We train GMM-DAE multiple times for this dataset, where each time only data from one scene is used for training. 

\subsection{Implementation Details}
We use YOLOv3 with the default weights trained on the Microsoft COCO dataset. We use a frame size of $384\times384$ and a confidence threshold of 0.3 as YOLOv3's hyperparameter values. We use all class labels available in YOLOv3 to avoid any detection bias, but only use those detected objects relevant to each dataset. When computing the dynamic images, we set the time stride to $t=10$. We use a standard deviation $\sigma=0.01$ for generating Gaussian noise and corrupt the data in the DAEs. We train the DAEs with a initial learning rate of 0.01, learning rate decay, a batch size of 1000 and $\beta=0.0001$ for $\ell_2$-regularization. We use the Adam optimizer \cite{Adam} with no more than 100 training epochs, where the specific number of epochs depends on the amount of training data. We recommend using $k=15$ Gaussian clusters to estimate the distribution of latent representations. As scenes may vary in terms of abnormal event types, it is recommended to use a simple grid search to get an optimal parameter set $\{\lambda_1, \lambda_2, \lambda_3, \lambda_4\}$ for late fusion.

\subsection{Comparison with SOTA models}
We calculate the normalized anomaly score of the current frame $I^t$ as follows:

\begin{equation}
\centering
S(I^t) = \frac{A(I^t) - \min_T A(I^T)}{\max_T A(I^T) - \min_T A(I^T)} \in [0, 1],
\label{eq:AnoScore}
\end{equation}

\noindent where $\max_T A(I^T)$ and $\min_T A(I^T)$ are the highest and the lowest anomaly scores among all frames of the same scene, respectively. We use the frame-level Area Under the Receiver Operating Characteristics (AUROC) curve, in \%, as the evaluation metric, which reflects how a model performs in terms of true detections and false detections. The AUROC values of our model and several SOTA models on the three datasets are tabulated in Table \ref{table:AUROCcomparison}. We highlight the results of our model and those of the best performing models.

\setlength{\tabcolsep}{4pt}
\begin{table}
\begin{center}
\caption{Frame-level AUROC values (\%) of several video anomaly detection models on the UCSD Ped2, CUHK Avenue and ShanghaiTech datasets.}
\vspace*{2mm}
\label{table:AUROCcomparison}
\begin{tabular}{|l|l|l|l|}
\hline
Model & UCSD Ped2 & CUHK Avenue & ShanghaiTech\\
\hline\hline
MPPCA \cite{MPPCA} & 69.3 & - & -\\
MPPCA+SFA \cite{UCSD} & 61.3 & - & -\\
MDT \cite{UCSD} & 82.9 & - & -\\
Unmasking \cite{UnMasking} & 82.2 & 80.6 & -\\

AMDN \cite{Xu} & 90.8 & - & -\\
FRCN action \cite{DGK} & 92.2 & \textbf{89.8} & -\\
Conv-AE \cite{3DConv2StreamR} & 90.0 & 70.2 & 60.9\\
STAE \cite{3DConvRP} & 91.2 & 77.1 & -\\
GANs \cite{GANROIOF} & 93.5 & - & -\\

FFP+MC \cite{GANNB} & 95.4 & 85.1 & 72.8\\
LSA \cite{LSA} & 95.4 & - & 72.5\\
MLAD \cite{GANMLP} & \textbf{99.21} & 71.54 & -\\
sRNN-AE \cite{sRNN-AE} & 92.21 & 83.48 & 69.63\\
MemAE \cite{MemAE} & 94.1 & 83.3 & 71.2\\
UnetGAN \cite{U-NetGAN} & 96.2 & 86.9 & -\\
MLEP-FP \cite{MLEP} & - & 89.2 & 73.4\\
MemAE2020 \cite{MemAE2020} & 97.0 & 88.5 & 70.5\\
SDOR \cite{SDOR} & 83.2 & - & -\\
SAGC \cite{SAGC} & - & - & 76.1\\

\hline
GMM-DAE & \textbf{96.5} & \textbf{89.3} & \textbf{81.2}\\
\hline
\end{tabular}
\end{center}
\end{table}
\setlength{\tabcolsep}{1.4pt}

As shown in Table \ref{table:AUROCcomparison}, GMM-DAE attains SOTA performance on the ShanghaiTech dataset (81.2\%), which is the most challenging dataset; namely, GMM-DAE improves the best performing model by 5.1\%. It is important to note that although the model in \cite{GraphNoiseCleaner} is reported to achieve 84.4\% AUROC value on the ShanghaiTech dataset, it uses a different split scheme on the test data. This scheme is unique to models that apply positive/negative video bags. Hence, \cite{GraphNoiseCleaner} does not report performance comparisons with other major SOTA models as it is unfair to carry out such comparisons. Thus, we omit the model in \cite{GraphNoiseCleaner} in Table \ref{table:AUROCcomparison}.

GMM-DAE achieves competitive performance on the UCSD Ped2 and CUHK Avenue datasets. We notice that MLAD \cite{GANMLP} and FRCN action \cite{DGK} attain SOTA results on these two datasets, respectively, but they do not obtain balanced and outstanding performance on all three datasets; e.g., MLAD performs very well on the UCSD Ped2 dataset (99.21\%) but poorly on the CUHK Avenue dataset (71.54\%), and has not been tested on the ShanghaiTech dataset by its authors. In comparison, our model attains very competitive results on these datasets. i.e., 96.5\% and 89.3\% on the UCSD Ped2 and CUHK Avenue datasets, respectively, which shows the robustness of our model to different scenes.

\begin{figure*}[!htb]
\centering
\subfloat{\includegraphics[width=0.32\textwidth]{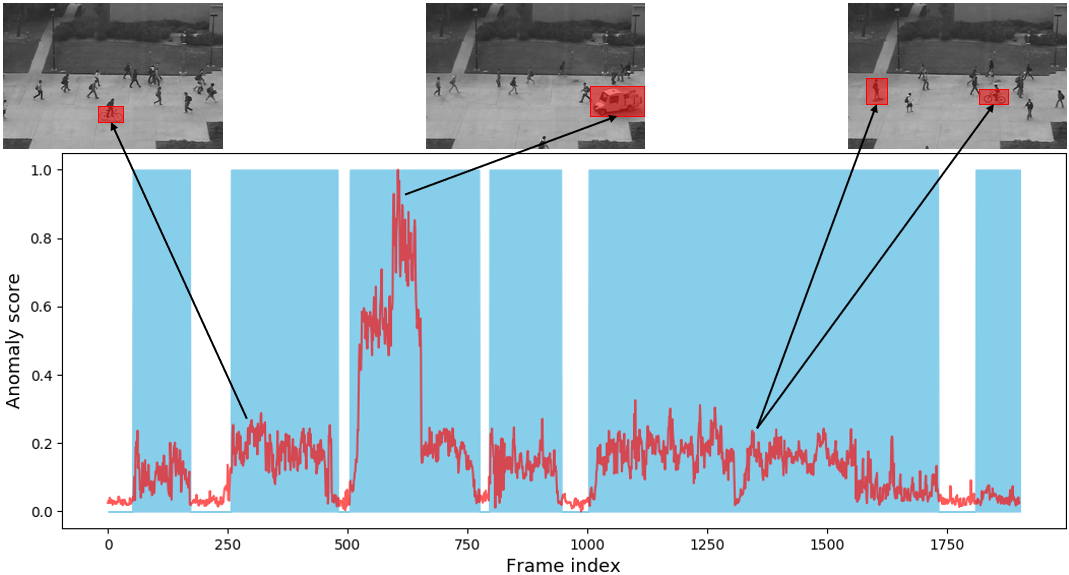}}
\hfill
\subfloat{\includegraphics[width=0.32\textwidth]{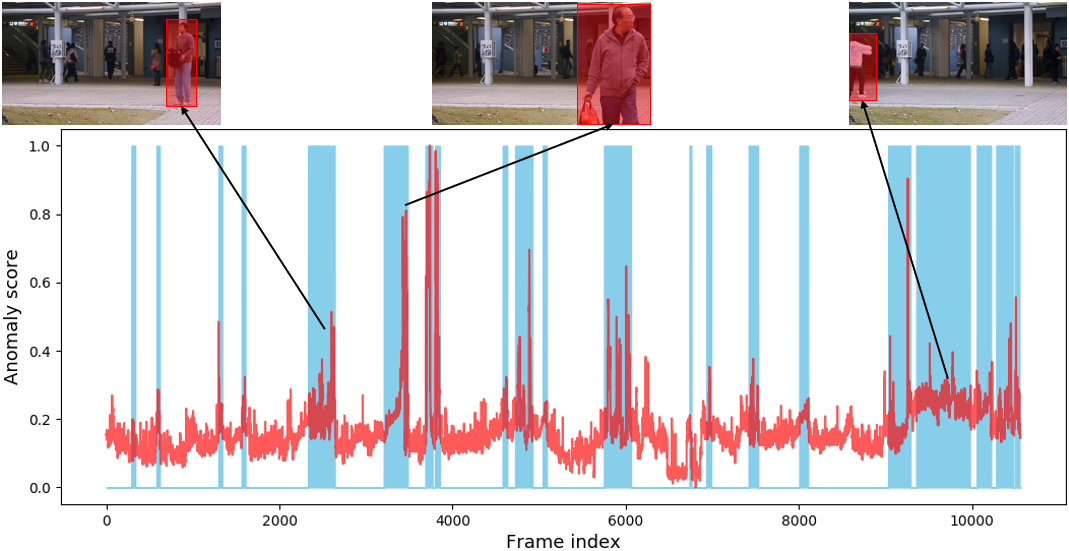}}
\hfill
\subfloat{\includegraphics[width=0.32\textwidth]{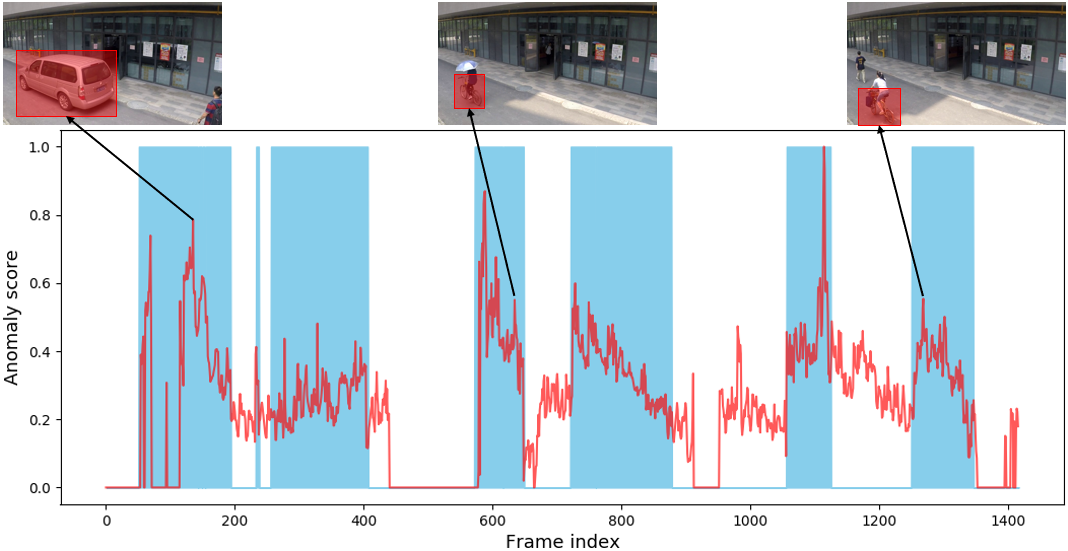}} \\
\subfloat{\includegraphics[width=0.19\textwidth]{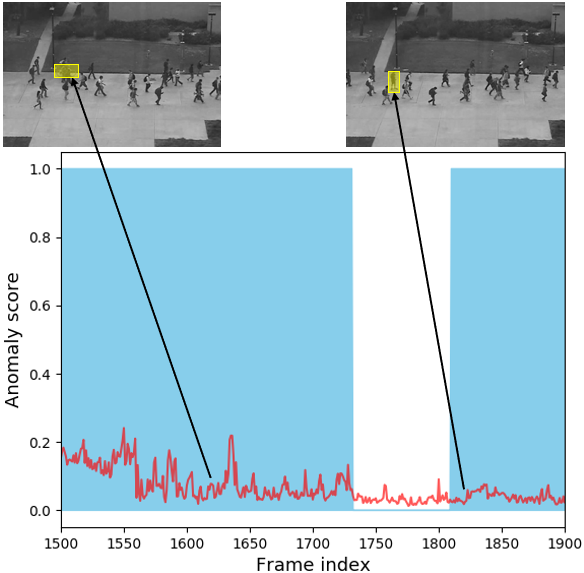}}
\hspace{0.35cm}
\subfloat{\includegraphics[width=0.38\textwidth]{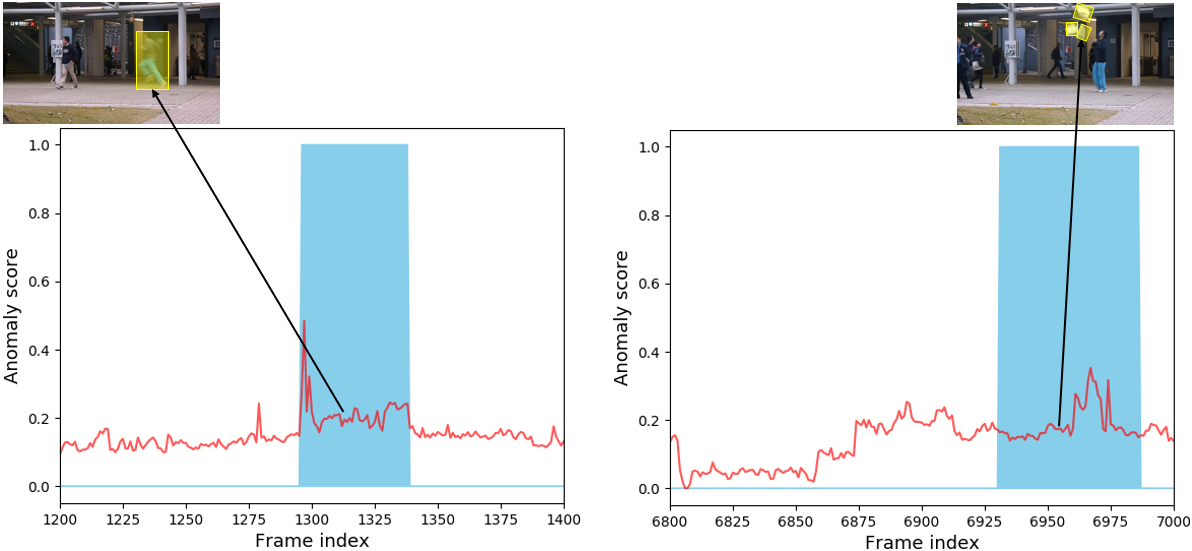}}
\hspace{0.35cm}
\subfloat{\includegraphics[width=0.19\textwidth]{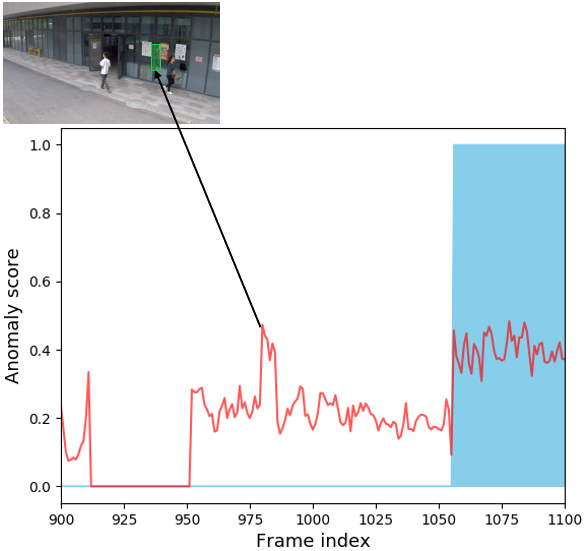}}
\caption{Performance analysis. Blue areas indicate abnormal frames. Red boxes indicate correct detections. Yellow boxes indicate false negative detections. Green boxes indicate false positive detections. First row: abnormality distribution of all frames of the UCSD Ped2 (left), CUHK Avenue (middle) and ShanghaiTech-Scene006 (right) datasets, together with examples of true positive detections. Second row: abnormality distribution of representative false negative detections on the UCSD Ped2 (leftmost) and CUHK Avenue (middle-left \& middle-right) datasets, and false positive detections on the ShanghaiTech-Scene006 dataset (rightmost). Activities from left to right in the first row: cycling, vehicle moving, skateboard riding \& cycling (left); backpack throwing, trespassing, dancing (middle); vehicle moving, cycling, cycling (right). Activities from left to right in the second row: cycling with occlusion, slowly skateboard riding (leftmost); sprinting, paper throwing (middle-left \& middle-right); mirror reflection of an individual walking (rightmost).}
\label{fig:PerformanceAnalysis}
\end{figure*}

\subsection{Performance Analysis}
In this section, we thoroughly analyze the performance of GMM-DAE to highlight its strengths and areas for improvement. We plot the abnormality distributions of the three evaluated datasets, as well as several representative true positive, false negative, and false positive detections in Fig.~\ref{fig:PerformanceAnalysis}.

As shown in Fig.~\ref{fig:PerformanceAnalysis}, when testing GMM-DAE on the UCSD Ped2 dataset, the model performs generally well in terms of minimizing false positive detections because the majority of high anomaly scores tend to happen within the blue areas. GMM-DAE can then easily detect abnormal events with higher anomaly scores; e.g., the vehicle moving, which triggers high anomaly scores in this case. We observe that most false negative detections in this case are due to occlusions, e.g., occlusions in the cyclists, or objects with a very slow motion pattern similar to the motion of pedestrians, e.g., those riding a skateboard at a slow speed. Moreover, GMM-DAE also minimizes false positive detections on the CUHK Avenue dataset. As for the false negative detections, the model fails at detecting extremely fast moving objects (a person sprinting) and small papers thrown in the air. In Scene006 of the ShanghaiTech dataset, GMM-DAE produces a false positive detection as indicated by the green box on the rightmost sub-figure in the second row. This detection is a reflection of an individual on a mirror. Our model regards this reflection as an abnormal object because the model has never seen the texture of such a reflection before.

Overall, GMM-DAE performs very well in terms of false positive detections in the majority of cases. It is important to acknowledge that YOLOv3 has deficiencies in detecting objects with occlusions, extremely fast moving objects and those with unusual shapes, which may lead to false negative detections. Our model uses approximate rank pooling to produce dynamic images to learn motion representations. Such dynamic images not be informative enough when objects move slowly, which may produce false negative detections, e.g., a slow-moving skateboard rider in the UCSD Ped2 dataset. Interestingly, YOLOv3 correctly detects mirror reflections as actual objects in the ShanghaiTech dataset, which makes the model produce several false positive detections due to the relatively weak discriminative power on normal, but non-real, objects.

\subsection{Discussions}

\setlength{\tabcolsep}{4pt}
\begin{table}
\begin{center}
\caption{Frame-level AUROC values (\%) on the UCSD Ped2 dataset using different components of the proposed GMM-DAE model.}
\vspace*{2mm}
\label{table:AblationStudy}
\begin{tabular}{cc} \toprule
Method & AUROC\\ \midrule
DAE+OP & 90.1\\
DAE+OP+DP & 93.9\\
DAE+OP+OL & 94.2\\
DAE+OP+OL+DP+DL (GMM-DAE)& \textbf{96.5}\\
AE+OP+OL+DP+DL & 95.8\\ \bottomrule
\end{tabular}
\end{center}
\end{table}
\setlength{\tabcolsep}{1.4pt}

\begin{figure*}[!t]
\centering
\subfloat{\includegraphics[width=0.35\textwidth]{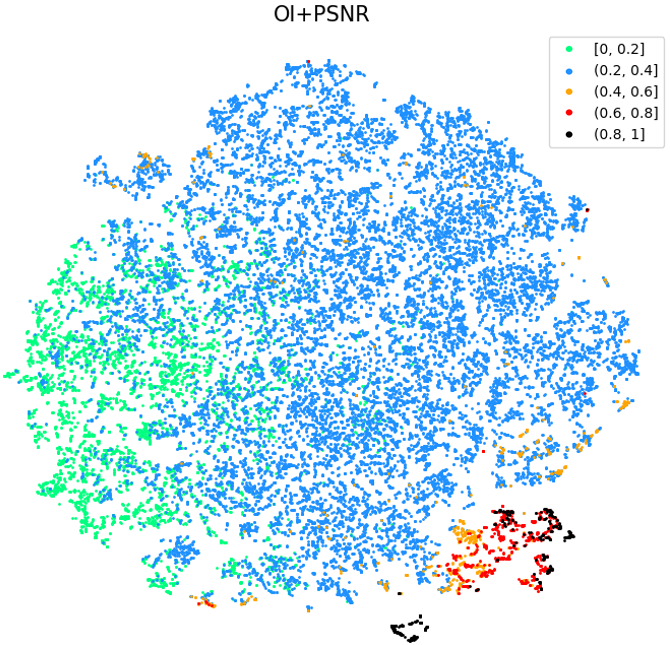}}
\hspace{2cm}
\subfloat{\includegraphics[width=0.35\textwidth]{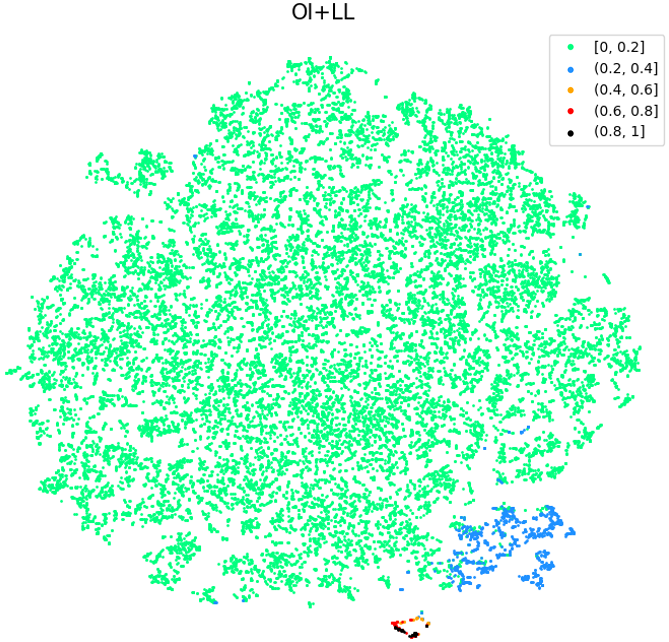}}
\caption{Distribution of the generated manifolds based on the normalized anomaly scores using PSNR values (left: OI+PSNR) and latent likelihood values (right: OI+LL) on the UCSD Ped2 test videos. Each color represents a range of anomaly scores.}
\label{fig:T-sne}
\end{figure*}

\noindent \textbf{Ablation Study}. In order to understand the importance of the various components of GMM-DAE, we carry out an ablation study on the UCSD Ped2 dataset on which GMM-DAE performs generally well. The results of this study are summarized in Table \ref{table:AblationStudy}. In this table, OP and OL indicate using patches along the appearance pipeline with the PSNR and the latent likelihood values, respectively, as the criterion to detect anomalies. DP and DL indicate using dynamic images along the motion pipeline with the PSNR and the latent likelihood values, respectively, as the criterion to detect anomalies. Finally, DAE and AE indicate, respectively, using a DAE or a vanilla AE for both pipelines. Note that DAE+OP+DP and DAE+OP+OL outperform DAE+OP by $3.8\%$ and $4.1\%$, respectively. This confirms that combining appearance and motion information, as well as modelling latent distributions, are essential for accurately detecting video anomalies. The best performance ($96.5\%$) is attained when all components of GMM-DAE are used, i.e., DAE+OP+OL+DP+DL. It is well-known that DAEs tend to outperform vanilla AEs w.r.t. informative feature extraction. This is confirmed in the results attained by AE+OP+OL+DP+DL. Namely, when using a vanilla AE in both pipelines, the performance of GMM-DAE drops by $0.7\%$.

\noindent \textbf{Latent Space Visualization}. 
In order to understand the learned probability distributions of the generated latent manifolds and their importance, we use T-distributed Stochastic Neighbor Embedding (T-SNE) to visualize the generated distribution of the patches (OI) of the UCSD Ped2 test videos \cite{T-SNE,T-SNEacc}. Fig.~\ref{fig:T-sne} plots the learned distributions based on the normalized anomaly scores using PSNR values (OI+PSNR) and latent likelihood values (OI+LL), where the color map comprises five colors to indicate five distinct ranges of anomaly scores. In this figure, we can see that manifolds with high anomaly scores (red and black points) are more closely distributed than those with low anomaly scores (green and blue points), which tend to be randomly distributed. It can also be seen that texture can be used to distinguish between manifolds with high and low anomaly scores (see OI+PSNR plot). This shows that data reconstruction is a useful criterion to detect abnormalities in less challenging dataset. However, comparing the two plots in Fig. ~\ref{fig:T-sne}, one can clearly see that the OI+PSNR plot depicts a distribution with more mixed-up colors and more high-anomaly manifolds, while the OI+LL plots depicts the opposite situation with clear clusters. In fact, for the UCSD Ped2 test videos, only a small portion of patches actually depict abnormal events, thus the data reconstruction criterion, i.e., PSNR, has important limitations, as it may be an inaccurate solution to video anomaly detection especially for challenging datasets. Moreover, the OI+LL plots depicts many low-anomaly manifolds that are depicted as having a high anomaly score in the OI+PSNR plot. This further indicates that the data reconstruction criterion can incorrectly result in high anomaly scores due to the high MSE values obtained even for normal inputs \cite{MemAE}. By relaying on both, PSNR and latent likelihood values, GMM-DAE learns appropriate Gaussian blobs from training data to accurately detect abnormal data in the latent space.

\section{Conclusion} \label{Conclusion}
We proposed a deep probabilistic model (GMM-DAE) for video anomaly detection. GMM-DAE detects abnormal patterns by relying on PSNR values from data reconstruction and likelihood values of low-dimensional representations estimated by learned GMMs. Hence, GMM-DAE probabilistically solves video anomaly detection as an unsupervised outlier detection task. We showed that GMM-DAE achieves competitive performance on the UCSD Ped2 and Avenue datasets, while achieving SOTA performance on the ShanghaiTech dataset. We also conducted a detailed performance analysis, which has shown that the performance of GMM-DAE is restricted by the accuracy of the object detector used. Our future work will mainly focus on novel density estimation and spatio-temporal representation learning methods that work without an object detector.

\printbibliography

\end{document}